\def\year{2020}\relax
\documentclass[letterpaper]{article} 
\usepackage{aaai20}
\usepackage{times}  
\usepackage{helvet} 
\usepackage{courier}  
\usepackage[hyphens]{url}  
\usepackage{graphicx} 
\usepackage{subcaption}
\usepackage{amsmath}
\usepackage{amssymb}
\usepackage{bbm}
\usepackage{booktabs}
\usepackage{color}
\usepackage{xcolor}
\usepackage{enumitem}
\usepackage{colortbl}
\usepackage{comment}
\usepackage{morefloats} 
\usepackage[none]{hyphenat}
\usepackage{etoolbox,capt-of}
\makeatletter
\newcommand\footnoteref[1]{\protected@xdef\@thefnmark{\ref{#1}}\@footnotemark}
\makeatother

\DeclareMathOperator*{\argmax}{arg\,max}

\newcommand{\doubleunderline}[1]{\underline{\underline{#1}}}

\usepackage{array}
\newcolumntype{P}[1]{>{\raggedright\arraybackslash}p{#1}}

\title{Characterizing Collective Attention via Descriptor Context: A Case Study of Public Discussions of Crisis Events}
\author{Ian Stewart \: 
Diyi Yang \:
Jacob Eisenstein \\
Georgia Institute of Technology \\ School of Interactive Computing \\ Atlanta, GA 30308 \\ \texttt{istewart6@gatech.edu, diyi.yang@cc.gatech.edu, jacobe@gatech.edu}
}

\begin{document}

\maketitle

\begin{abstract}
Social media datasets make it possible to rapidly quantify \emph{collective attention} to emerging topics and breaking news, such as crisis events.
Collective attention is typically measured by aggregate counts, such as the number of posts that mention a name or hashtag.
But according to rationalist models of natural language communication, the collective salience of each entity will be expressed not only in how often it is mentioned, but in the form that those mentions take.
This is because natural language communication is premised on (and customized to) the expectations that speakers and writers have about how their messages will be interpreted by the intended audience.
We test this idea by conducting a large-scale analysis of public online discussions of breaking news events on Facebook and Twitter, focusing on five recent crisis events.
We examine how people refer to locations, focusing specifically on contextual descriptors, such as ``\emph{San Juan}'' versus ``\emph{San Juan, Puerto Rico}.'' 
Rationalist accounts of natural language communication predict that such descriptors will be unnecessary (and therefore omitted) when the named entity is expected to have high prior salience to the reader. 
We find that the use of contextual descriptors is indeed associated with proxies for social and informational expectations, including macro-level factors like the location's global salience and micro-level factors like audience engagement.
We also find a consistent decrease in descriptor context use over the lifespan of each crisis event.
These findings provide evidence about how social media users communicate with their audiences, and point towards more fine-grained models of collective attention that may help researchers and crisis response organizations to better understand public perception of unfolding crisis events.
\end{abstract}
\section{Introduction}

Breaking news events, such as crises, can attract significant \emph{collective attention} from the general public~\cite{lin2014}, resulting in bursts of discussion on social media~\cite{leavitt2014,lehmann2012}.
During such events, attention is often focused on entities (e.g., people, locations, and organizations) that are highly relevant to the unfolding event~\cite{wakamiya2015}.
A spike in attention directed toward a particular entity may signal an important update, such as the need for aid for the location~\cite{varga2013}.

While collective attention is often measured with activity metrics such as post volume~\cite{mitra2016}, we propose to characterize collective attention not by \emph{how often} an entity is mentioned, but by \emph{how} it is mentioned.
Content-based metrics can be more accurate, as they are less sensitive to data sparsity and biases, such as when a crisis limits internet accessibility.
Furthermore, measuring the content of collective attention can provide insight into writer's expectations of reader knowledge, which may become visible as they adapt their writing to address a local or general audience.

\begin{figure*}
    \begin{subfigure}{0.49\textwidth}
    \centering
    \includegraphics[width=\textwidth]{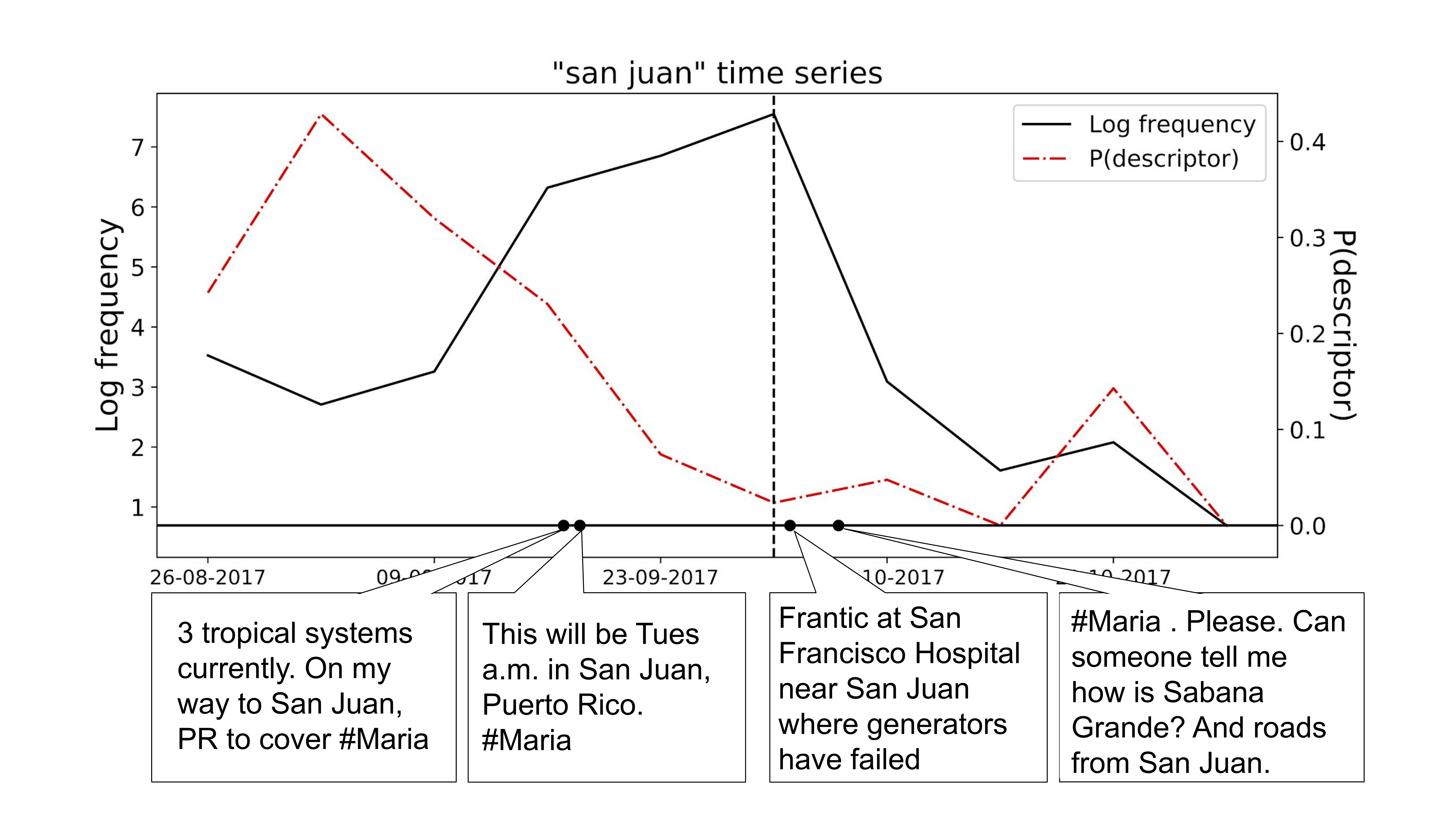}
    \caption{Timeline for mentions of ``\emph{San Juan}'' during Hurricane \\ Maria, with example tweets below.}
    \label{fig:example_plot_san_juan}
    \end{subfigure}
    \begin{subfigure}{0.49\textwidth}
    \centering
    \includegraphics[width=\textwidth]{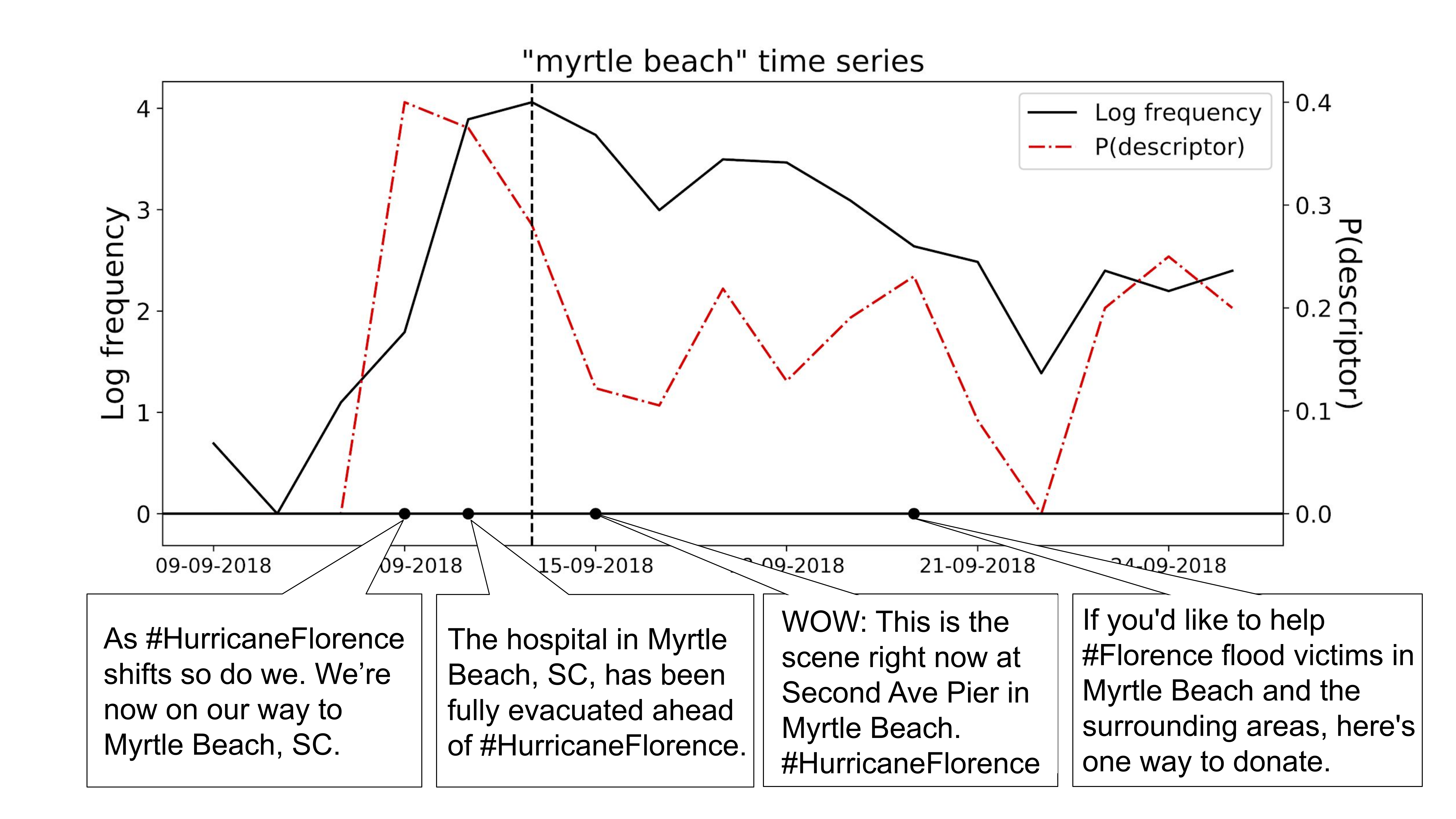}
    \caption{Timeline for mentions of ``\emph{Myrtle Beach}'' during Hurricane \\ Florence, with example tweets below.}
    \label{fig:example_plot_myrtle_beach}
    \end{subfigure}
    \caption{Example of collective attention expressed toward location mentions in discussion of various hurricanes on Twitter.
    Left y-axis (black solid line) indicates the location's log frequency, right y-axis (red dotted line) indicates the location's probability of receiving a descriptor phrase such as ``\emph{San Juan, \underline{Puerto Rico}}.''
    For example, a 25\% probability for ``\emph{San Juan}'' means that 25\% of all mentions of ``\emph{San Juan}'' had a descriptor phrase.}
    \label{fig:example_entity_freq_descriptor_plot}
\end{figure*}

To understand how the content of entity references can change with collective attention, consider Hurricane Maria, which struck Puerto Rico in September 2017. 
As more people became familiar with the locations mentioned in news coverage about the island~\cite{dijulio2017}, news headlines and articles referred to ``\emph{San Juan}'' without extra contextual descriptors such as ``\emph{the capital of Puerto Rico}.''
This is consistent with a rational model of communication in which linguistic contextualization is used for entities that might otherwise be unknown or ambiguous~\cite{prince1992,staliunaite2018}: as San Juan became increasingly salient through repeated mentions, readers could be expected to understand the reference without additional context.
Figure~\ref{fig:example_entity_freq_descriptor_plot} presents evidence from Twitter in favor of this hypothesis: the locations of San Juan (\ref{fig:example_plot_san_juan}) and Myrtle Beach (\ref{fig:example_plot_myrtle_beach}) received fewer contextualizing descriptors following peaks in the volume of mentions in discussions of Hurricanes Maria and Florence respectively.

While global salience plays an important role, more fine-grained factors are also at work. 
Authors may add or remove context information based on their expectations about their specific audience and on the availability of additional context such as hyperlinks to external articles.
Furthermore, even if we observe an aggregate change in collective attention, we cannot be sure whether the trend is due to change in the author population or a change in the behavior of individual authors.
We disentangle these macro-level and micro-level factors in a multivariate analysis of descriptor phrase usage in discussions of crisis events, using data from public discussions of five recent natural disasters on Facebook and Twitter.
We investigate the usage of contextual descriptor phrases in references to locations affected by hurricanes, which we link to various proxies for information expectations: temporal trends in relation to the event itself; properties of the author, audience, and entity; and the presence of extra-linguistic context such as hyperlinks and images. This investigation addresses the following research questions:
\begin{itemize}
    \item \textbf{RQ1}: What factors influence the use of descriptor context in reference to locations of hurricane events? 
    \item \textbf{RQ2a}: How does the use of descriptor context for locations change over time at a collective level?
    \item \textbf{RQ2b}: How does the use of descriptor context for locations change at an individual author level? 
    
\end{itemize}

We now briefly summarize the high-level findings. In a dataset of Facebook posts from public groups concerning Hurricane Maria relief, we find that location mentions received descriptors more often when the locations were not local to the group of discussion, suggesting that descriptors may be used to help explain new information to audiences.
In a dataset of public Twitter posts related to five hurricane events, we find that the aggregate rate of descriptor phrases decreased following the peaks in these locations' collective attention, supporting prior findings on named entity references in professional newstext~\cite{staliunaite2018}.

However, this result is supplemented by more fine-grained effects that also support a rational account of entity reference: authors used fewer descriptors if they had mentioned a location before, and more descriptors if their previous posts received high audience engagement (as measured by retweets and likes), which suggest a larger (and potentially less contextualized) audience.
In sum, this work identifies strong connections between shared information and entity references, which supplements existing linguistic theory while offering researchers and practitioners new tools for measuring and understanding collective attention in crisis events.
All code is available for use.\footnote{Repository here: \\ \url{https://github.com/ianbstewart/collective_attention}.}
\section{Related Work}
The term \emph{collective attention} refers to the degree to which public interest and awareness is focused on individual events, entities, or topics~\cite{sasahara2013}.
Collective attention can shift either rapidly or gradually~\cite{wu2007}, often in response to large-scale events such as sports games~\cite{lehmann2012}, natural disasters~\cite{varga2013}, and political controversy~\cite{garimella2017}.
Prior studies of social media have often quantified collective attention using the volume of posting and sharing activity~\cite{leavitt2014,mitra2016}.
We supplement this prior work by focusing on how linguistic content changes with collective attention over crisis events.

\begin{table*}[]
 \centering
 \small
 \begin{tabular}{l P{2.5cm} r r r l}
 Event & Hashtags & Date range & Tweets & LOCATION NEs & LOCATION examples \\ \toprule
Florence & \#florence, \#hurricaneflorence & {[}30-08-18, 26-09-18] & 66595 & 28670 & Wilmington, New Bern, Myrtle Beach \\ 
Harvey & \#harvey, \#hurricaneharvey & {[}17-08-17, 10-09-17] & 679400 & 181636 & Houston, Corpus Christi, Rockport \\ 
Irma & \#irma, \#hurricaneirma & {[}29-08-17, 20-09-17] & 809423 & 229315 & Miami, Tampa, Naples \\ 
Maria & \#maria, \#hurricanemaria, \#huracanmaria & {[}15-09-17, 09-10-17] & 313088 & 57237 & San Juan, Vieques, Ponce \\ 
Michael & \#michael, \#hurricanemichael & {[}06-10-18, 23-10-18] & 52506 & 22007 & Panama City, Mexico Beach, Tallahassee 
 \end{tabular}
 \caption{Summary statistics for Twitter data. }
 \label{tab:twitter_data_table}
\end{table*}

When referring to an entity such as a location, the writer may add descriptive information in the form of a dependent clause~\cite{kang2019}. The dependent clause may describe attributes of the entity that are relevant to a specific topic, such as ``\emph{San Juan, epicenter of the Hurricane Maria relief effort},'' or attributes that are generally relevant, such as ``\emph{San Juan, Puerto Rico}.''
Rationalist models of communication~\cite{grice1975logic,prince1992} predict that such descriptors will be particularly necessary for entities that are not already salient to the audience. 
These predictions have received empirical support in corpus analyses: \citeauthor{siddharthan2011} (\citeyear{siddharthan2011}) found that highly \emph{salient} or \emph{important} named entities in a summary of a news story are more likely to be understood as shared knowledge and are therefore less likely to need a descriptor phrase.
Similarly, a diachronic analysis of newspaper articles showed that entities tended to receive fewer descriptors as they became more familiar over time~\cite{staliunaite2018}.
In laboratory settings, speakers have been found to modulate their use of contextualizing information depending on the expected entity salience to \emph{specific} listeners~\cite{galati2010}, but such fine-grained analyses have not previously been explored with corpus data.
This study contributes to this literature by extending the focus from professional print media and laboratory experiments to social media, where it is possible to measure fine-grained social and informational factors in naturally-occurring text.

We choose crisis events as the domain of study because they present a natural example of wide-scale information sharing with respect to a shared event~\cite{houston2015}.
During crisis events, citizens often turn to social media to share their first-hand experiences~\cite{soden2018} and to seek information~\cite{varga2013}. Non-governmental organizations (NGOs), government agencies, and media outlets often rely on social media as a barometer for the experiences of people affected by crises~\cite{imran2015}.
These organizations' relief efforts may be informed by the collective attention directed toward actionable needs of crisis victims~\cite{palen2018}. 
Prior work has proposed a variety of computational methods to extract actionable information from social media for crisis responders~\cite{olteanu2015,temnikova2015}, such as locations that need aid.

Our work proposes a fine-grained metric for collective attention to provide insight into how authors frame their writing to address audience expectations during crisis events, which can help organizations understand information needs among the public~\cite{murthy2017}.
We discuss further implications for crisis informatics in \S~\ref{subsec:implications}.
\section{Data}

Crisis events present a useful case study for the development of collective attention, due to the large volume of online participation and uncertainty among event observers towards the situation~\cite{varga2013}.
We chose to study the collective attention changes in public discourse related to hurricanes, due to hurricanes' lasting economic impact, their broad coverage in the news, and their relevance to specific geographic regions.
We collected social media data related to five recent hurricanes. 
The remainder of this section describes the data collection (\S~\ref{sec:data_collection}), location detection (\S~\ref{sec:detect_filter_locations}), and descriptor detection (\S~\ref{sec:detect_context}) for the following datasets:
\begin{enumerate}
    \item Twitter: 2 million public tweets related to 5 major hurricanes, collected in 2017 and 2018.
    \item Facebook: 30,000 posts from 60 public groups related to disaster relief in Hurricane Maria, collected in 2017.
\end{enumerate}

\begin{table}[]
    \centering
    \small
    \begin{tabular}{l r r r}
        Event & Authors & Tweets & LOCATION NEs\\ \toprule
        Florence & 186 & 17624 & 29066 \\ 
        Harvey & 164 & 31563 & 50050 \\ 
        Irma & 178 & 45913 & 77114 \\ 
        Maria & 139 & 11332 & 18204 \\ 
        Michael & 146 & 8828 & 14655  \\
    \end{tabular}
    \caption{Summary statistics for active authors on Twitter.}
    \label{tab:active_author_data_table}
\end{table}
\begin{table*}[]
    \small
    \centering
    \begin{tabular}{l P{3cm} l}
        Phrase patterns & Dependency types & Example \\ \toprule
        \underline{LOCATION} + \doubleunderline{LOCATION\_STATE} & n/a & \underline{San Juan}, \doubleunderline{PR} \\
        \underline{LOCATION} + $\text{[\doubleunderline{LOCATION\_CONTEXT}]}_{\text{MODIFIER}}$ & adjective, apposition, preposition, numeric modifier & \underline{San Juan}, [capital of \doubleunderline{Puerto Rico}] \\
        $[\underline{\text{LOCATION}} \: + \: \doubleunderline{\text{LOCATION\_CONTEXT}}]_{\text{NOUN\_COMPOUND}}$ & nominal, compound, apposition & the [\underline{Vega Alta} neighborhood of \doubleunderline{San Juan}] \\
        \underline{LOCATION} + $[\doubleunderline{\text{LOCATION\_STATE}}]_{\text{CONJUNCTION}}$ & conjunction & \underline{San Juan}, Guayama [and Vieques, \doubleunderline{Puerto Rico}] \\
    \end{tabular}
    \caption{Phrase patterns to capture descriptor phrases in location mentions. Head location marked with underline, context location marked with double underline.}
    \label{tab:descriptor_phrase_patterns}
\end{table*}
\subsection{Collection}
\label{sec:data_collection}
\paragraph{Twitter Dataset}
The Twitter posts were collected using hashtags from five major hurricanes: Hurricane Florence (2018), Hurricane Harvey (2017), Hurricane Irma (2017), Hurricane Maria (2017), and Hurricane Michael (2018).
We used hashtags that contained the name of the event in full and shortened form, e.g. \#Harvey and \#HurricaneHarvey for Hurricane Harvey.

During 2017 and 2018, we streamed tweets that contained hashtags related to the natural disasters at the start of each disaster for up to one week after the dissipation of the hurricane.\footnote{Dates are based on estimates from the National Oceanic and Atmospheric Administration (NOAA). For example, estimates for Hurricane Harvey are available
at \url{https://www.nhc.noaa.gov/data/tcr/AL092017_Harvey.pdf} (accessed January 2019).}
We augmented this data with additional tweets available in a 1\% Twitter sample that contains the related hashtags, restricting our time frame to one day before the formation of the hurricane and one week after the dissipation of the hurricane.
Manual inspection revealed minimal noise generated by the inclusion of the name-only hashtags (e.g., \#Harvey).

Summary statistics of the Twitter data are presented in Table~\ref{tab:twitter_data_table}.
We also collected additional event-related tweets from the most frequently-posting authors in each dataset (``active authors''), which were needed to evaluate per-author changes (RQ2b; see \S~\ref{sec:descriptor_change_individual}). 
Table~\ref{tab:active_author_data_table} summarizes the detailed statistics about the active author data.

\paragraph{Facebook Dataset} 
The Facebook data was collected in the aftermath of Hurricane Maria by searching for public discussion groups that included at least one of Puerto Rico's municipalities in the title (e.g. ``\emph{Guayama: Hurac\'an Maria}'' refers to Guayama municipality).
Relatives and friends of Puerto Ricans often posted in these groups to seek additional information about those still on Puerto Rico, who could not be reached by telephone due to infrastructure damage.
We restricted our analysis to Facebook groups related to Hurricane Maria because the limited information available caused more discussion of specific locations, as compared to the other hurricane events that had more up-to-date information available online.

In total, we collected 31,414 public posts from 61 groups, from the time of their creation to one month afterward (September 20 to October 20, 2017).
Spanish was the majority language in these posts, so only posts in Spanish were retained, using \texttt{langid.py}~\cite{lui2012}.\footnote{\url{https://github.com/saffsd/langid.py} (accessed October 2017)}
Due to Facebook data restrictions and API changes, we were unable to collect posts in Facebook groups for the other four hurricane events.

\subsection{Extracting and Filtering Locations}
\label{sec:detect_filter_locations}
We extracted mentions of locations using two systems for named entity recognition (NER): for English, we used a system that was explicitly adapted to Twitter data~\cite{ritter2011}\footnote{\url{https://github.com/aritter/twitter_nlp} (accessed January 2019)}
and for Spanish, we used a general purpose named entity recognizer~\cite{finkel2005}.\footnote{\url{https://nlp.stanford.edu/software/stanford-ner-2018-10-16.zip} (accessed January 2019)}
These systems are freely accessible and widely used, and achieve reasonably competitive performance.\footnote{For location entities, the English tagger has a reported F1 of $74\%$~\cite{ritter2011}, and the Spanish tagger has a reported F1 of $58\%$~\cite{finkel2009}, but these figures are not directly comparable due to genre differences across datasets. 
Both English and Spanish are considered ``high resource" languages for natural language processing, with hundreds of thousands of tokens of labeled data for named entity recognition~\cite{hovy2006ontonotes,taule2008ancora}. 
The extension of this data acquisition pipeline to languages that lack substantial labeled data may pose a significant challenge~\cite{rahimi-etal-2019-massively}.}
We manually evaluated the performance of these NER systems on a sample of tweets (100 tagged LOCATIONs per dataset, 500 total) and found reasonable precision for the \texttt{LOCATION} tag (81-96\% across all datasets).

For this work, we focus on named entities that may require descriptor phrases, which include cities and counties.
We therefore restrict our analysis to named entities (NEs) that (1) are tagged as \texttt{LOCATION}, (2) can be found in the GeoNames ontology,\footnote{\url{http://download.geonames.org/export/dump/allCountries.zip} (accessed September 2017)} (3) map to cities or counties in the ontology, (4) map to affected locations in the ontology, based on their location occurring in the region affected by the event, and (5) are unambiguous within the region affected by the event.
For instance, the string ``\emph{San Juan}'' is a valid location for the Hurricane Maria tweets because the affected region contains an unambiguous match for the string, but it is not a valid location for the Hurricane Harvey tweets because the affected region does not contain an unambiguous match.

\subsection{Extracting Descriptor Phrases}
\label{sec:detect_context}

One way in which writer can introduce a new entity to a discourse (e.g., ``\emph{San Juan}'') is by linking it to a more well-known entity (e.g., ``\emph{Puerto  Rico}'') in a descriptor phrase.
To detect this phenomenon, we identified location mentions that had dependent clauses that referred to better-known locations, using population as a proxy.
The underlying assumption is that a more well-populated location is be more likely to be known to readers, and can therefore help describe the preceding location.
The frequency of such descriptor phrases is the main dependent variable in this research: we hypothesize that authors are more likely to use such descriptor phrases when they expect readers to treat the location as new information, and less likely to do so when the location is already salient.

\begin{table*}[t]
\small
\centering
\begin{tabular}{l l P{10cm}}
Factor & Variable & Description \\ \toprule
\textbf{Importance} & Prior location mentions & Frequency of location within the group or event \\ \midrule
\textbf{Author} & In-group posts & Posts that an author made within a group \\
 & In-event posts & Posts that an author made about an event (log-transformed) \\
 & In-event posts about location & Posts that an author made about an event that mention the location (log) \\
 & Organization & Whether the author is predicted to be an organization (based on metadata) \\
 & Local & Whether the author is predicted to be local to the event (based on self-reported location) \\ \midrule
\textbf{Audience} & Location is local to group & Whether the location exists within the group's associated region \\
 & Group size & Number of unique members who have posted in the group \\
 & Prior engagement & Mean normalized log-count of retweets and likes received by an author (in t-1) \\
 & Change in prior engagement & Change in prior engagement received by an author (between t-2 and t-1) \\ \midrule
 \textbf{Information} & Has URL & Whether the post contains a URL \\
 & Has image/video & Whether the post contains a URL with an associated image/video \\ \midrule
\textbf{Time} & Time since start & Days since first post about event \\
 & During peak & Whether post was written during peak of collective attention toward location \\
 & Post peak & Whether post was written at least 1 day after the peak of collective attention toward location
\end{tabular}
\caption{Summary of explanatory variables and corresponding metrics, used for descriptor phrase prediction.}
\label{tab:descriptor_explanatory_variables}
\end{table*}

To extract sentence structure from text, we used dependency parsing, which decomposes sentences into directed acyclic graphs connecting words and phrases~\cite{eisenstein2019introduction}.
Following \citeauthor{staliunaite2018} (\citeyear{staliunaite2018}), we defined a set of dependencies to capture the \texttt{MODIFIER} phrase type in a subclause (adjectival clause, appositional modifier, prepositional modifier, numeric modifier) and another set of dependencies to capture the \texttt{COMPOUND} type in a super-clause (nominal modifier, compound, appositional modifier).
Table \ref{tab:descriptor_phrase_patterns} presents a summary of the phrase patterns that were used to capture descriptor phrases.
Taking into account the characteristics of text from two different domains, for the Twitter data we used the \texttt{spacy} shift-reduce parser~\cite{honnibal2015}\footnote{\url{https://spacy.io/usage} (accessed January 2019)}; for the Facebook data, the dependencies were extracted using the SyntaxNet transition-based parser~\cite{andor2016}.\footnote{\url{https://cloud.google.com/natural-language/docs/analyzing-syntax} (accessed January 2019)}
Our pilot experiments found that SyntaxNet achieved higher accuracy on Facebook posts, but we were unable to apply it to the larger Twitter dataset due to API restrictions.\footnote{As a robustness check we re-ran the analysis for the Facebook data using parses from \texttt{spacy}, and found relatively the same effect sizes for all variables considered (see \S~\ref{sec:descriptor_robustness_check} in the Appendix).}

\paragraph{Validation of Extraction Performance}
To assess the accuracy of our phrase patterns in capturing descriptor phrases, we asked two annotators (computer science graduate students) who had not seen the data to annotate a random sample of 50 tweets containing at least one location from each data set (250 tweets total).
The annotators received instructions on how to determine if a location was marked by a descriptor phrase, including examples that were not drawn from the data, and the annotators marked each location mention as either (1) a ``LOCATION + LOCATION\_STATE'' pattern, (2) one of the other descriptor patterns in Table \ref{tab:descriptor_phrase_patterns} or (3) no descriptor phrase.
The annotators achieved high agreement on each separate descriptor type (Cohen's $\kappa = 0.96$ for the state pattern, $\kappa = 0.91$ for the other patterns).
We then filtered posts with perfect agreement, ran dependency parsing on the posts and detected descriptor phrases using the phrase patterns proposed. 
We found that our phrase patterns achieved reasonable precision and recall (96.6\% and 87.5\% respectively) in identifying descriptor phrases compared to raters' annotations.
This validation check demonstrated that our proposed syntactic patterns can capture descriptor phrases reasonably well.
\section{Results}
We address our research questions in three analyses: static social factors, dynamic factors at the collective level, and dynamic factors at the individual level.  

\subsection{What Affects the Use of Descriptor Phrases?}
\label{sec:descriptor_local_test}
We first address RQ1, concerning which social factors influence the use of descriptor context when referring to locations of hurricane events.
We are particularly interested in indicators of whether locations may be considered shared knowledge within a community.
A descriptor phrase may be omitted for locations that are geographically local to a group of people, i.e. knowledge that already shared among the group and are therefore assumed to be \emph{old} information (e.g., if someone mentions the location ``\emph{San Juan}'' in a group based in a region containing San Juan).

\begin{table*}[h]
\small
\centering
\begin{tabular}{l P{3cm} | r r | r r | r r | r r}
 &  & \multicolumn{2}{l}{\textbf{RQ1} (Facebook)} & \multicolumn{2}{l}{\textbf{RQ1} (Twitter)} & \multicolumn{2}{l}{\textbf{RQ2a} (Twitter)} & \multicolumn{2}{l}{\textbf{RQ2b} (Twitter)} \\
Factor & Variable & Estimate & S.E. & Estimate & S.E. & Estimate & S.E. & Estimate & S.E. \\ \toprule
Intercept &  & -2.030 & 28.550 & -1.052* & 0.404 & -1.026* & 0.415 & -1.222 & 11.206 \\ \midrule
\textbf{Importance}  & Prior location mentions & -0.075 & 7.164 & -0.172* & 0.025 & -0.200* & 0.031 & -0.107 & 0.114 \\ 
\midrule
\textbf{Author}  & Author in-group posts & -0.328 & 0.522 & - & - & - & - & - & - \\
 & Author is organization & - & - & 0.093* & 0.033 & 0.092* & 0.035 & -0.149 & 0.115 \\
 & Author is local & - & - & -0.511* & 0.020 & -0.797* & 0.031 & -0.671* & 0.107 \\
 & Prior event-based posts (from author) & - & - & - & - & - & - & 0.110 & 0.093 \\
 & Prior location mentions (from author) & - & - & - & - & - & - & -0.237* & 0.091 \\ 
\midrule
\textbf{Audience}  & Local location & -0.623* & 0.106 & - & - & - & - & - & - \\
 & Group size & 0.121 & 0.040 & - & - & - & - & - & - \\
 & Prior engagement (author) & - & - & - & - & - & - & 0.292* & 0.052 \\
 & Change in prior engagement (author) & - & - & - & - & - & - & -0.004 & 0.042 \\ 
\midrule
\textbf{Information} & Has URL & - & - & -0.081* & 0.035 & -0.058 & 0.038 & -0.482* & 0.154 \\
 & Has image/video & - & - & 0.137* & 0.032 & 0.124* & 0.034 & 0.562* & 0.123 \\ 
\midrule
\textbf{Time}  & Time since start & - & - & - & - & -0.120* & 0.036 & -0.004 & 3.63 \\
 & During-peak & - & - & - & - & 0.004 & 0.038 & 0.144 & 0.122 \\
 & Post-peak & - & - & - & - & -0.127* & 0.049 & -0.189 & 0.157 \\ 
\midrule
Model deviance & ~ & \multicolumn{2}{r|}{469} & \multicolumn{2}{r|}{2954} & \multicolumn{2}{r|}{4127} & \multicolumn{2}{r}{2239} \\
Accuracy & ~ & \multicolumn{2}{r|}{71.3\%} & \multicolumn{2}{r|}{72.7\%} & \multicolumn{2}{r|}{73.3\%} & \multicolumn{2}{r}{75.0\%} \\
\end{tabular}
\caption{Logistic regression results for all analysis, predicting the presence of a descriptor phrase. 
All regressions include fixed effects for location, (for Facebook) group, and (for Twitter) event. 
* indicates $p < 0.05$, otherwise $p > 0.05$ after multiple hypothesis correction. 
All models have significantly higher log-likelihood as compared to the null model (see ``Model deviance'') and significantly higher accuracy than chance (see ``Accuracy''; accuracy computed over 10 runs of class-balanced sampled data).
Note that the author population for RQ2b is restricted to active authors and therefore different from the author population in RQ1 (Twitter) and RQ2a.}
\label{tab:descriptor_regression_all_results}
\end{table*}

We compared the rate of descriptor uses for location mentions in both Facebook and Twitter.
For the Facebook data, we determined whether the group's region contains the location mentioned based on whether the most likely match for the location in the gazetteer is contained in that region.\footnote{When a location string matches multiple location entities, we choose the one with the highest population.}
We also considered the following additional predictors: frequency with which the location is mentioned in prior posts (importance), author posting frequency in the group (author status), and group size (audience), as summarized in Table~\ref{tab:descriptor_explanatory_variables}.

For the Twitter data, we considered the folllowing predictors: location mention frequency in the Twitter sample (importance), whether the author is an organization or a local to the location (author status),\footnote{\label{social_status_footnote}See \S~\ref{sec:author_social_status} in the Appendix for details on determining whether an author is an organization or local.}, whether the post has a URL (information), and whether the post has an image or video (information).

We built separate logistic regression models for the Facebook and Twitter data. In both cases, the dependent variable is whether each location mention was accompanied by a descriptor phrase (N=18432 and N=49020, respectively).
In detail, we used an elastic net regression~\cite{zou2005}\footnote{An L2 regularization of 0.01 was chosen through grid search to maximize log-likelihood on held-out data (90-10 train/test split).} in order to reduce the risk of overfitting.
For this analysis, rare categorical values ($N<20$) for the fixed effects are replaced with \texttt{RARE} values to avoid overfitting to uncommon categories.
The columns ``\emph{RQ1 (Facebook)}'' and ``\emph{RQ1 (Twitter)}'' in
Table~\ref{tab:descriptor_regression_all_results} report the results of the logistic regression. 

On Facebook (see ``\emph{RQ 1 (Facebook)}'' in Table \ref{tab:descriptor_regression_all_results}), mentions of locations that are local to the group received significantly fewer contextual descriptors ($\beta=-0.623, p < 0.001$).
For example, in the group ``\emph{Hurricane Maria in Lajas}'' the mention of the municipality ``\emph{Lajas}'' does not receive an descriptor (``\emph{Do you know if the Bank is open in Lajas?}''), while in the group ``\emph{Guayama: Hurac\'{a}n Maria}'' the mention of ``\emph{Lajas}'' does receive an descriptor (``\emph{People who can bring water to Lajas Puerto Rico: they need water urgently}'').\footnote{Comments are translated from Spanish and paraphrased to preserve privacy.}
The other predictors did not have a statistically significant effect on descriptor use. 

On Twitter, there were several significant effects: more salient and important locations receive fewer descriptors ($\beta=-0.172, p < 0.001$); authors who are local to an event are less likely to include descriptor phrases ($\beta=-0.511, p < 0.001$), while organizational accounts on Twitter are more likely to use descriptor phrases ($\beta=0.093, p < 0.01$); in posts that contain URLs, descriptors are less likely to appear ($\beta=-0.081, p < 0.05$), but in posts that links to image or video, descriptors are more likely ($\beta=0.137, p < 0.001$). 

These findings are in accord with the view that authors customize their presentations based on the perceived information needs of readers. 
Additional context is unnecessary when writing for locals, or when writing about entities that are already salient, or have become salient through repeated mentions. 
Twitter accounts that represent large organizations are likely writing for large audiences who require more context; locals are more likely writing for their peers, who do not. 
Additional context can be provided by hyperlinks to detailed stories, but multimedia content such as images and videos do not serve the same purpose, and therefore require additional contextualization.

\subsection{Collective Change in Descriptor Context Use}
\label{sec:descriptor_change_aggregate}
We now turn to a temporal analysis of descriptor use, using longitudinal data from Twitter. 
As collective attention focuses on affected locations over the course of a crisis event, we those locations to require less contextualization. To test this theory, we augment the predictors from the previous section with two temporal variables: whether the message is posted during or after the peak volume in the discussion of the event, and how many days have elapsed since the start of the hurricane. 

The definition of the peak in collective attention is critical, because it determines the point at which an entity is expected to become shared knowledge in a discussion~\cite{staliunaite2018}.
Following \citeauthor{mitra2016} (\citeyear{mitra2016}), we defined the time of peak collective attention $\hat{t}_{i}$ for each location $i$ as the (24-hour) period during which it is mentioned the most frequently:
$\hat{t}_{i} = \argmax_{t \in T} f^{(i)}_{t}$,
where $f^{(i)}_{t}$ is the raw frequency of location $i$ at time $t$ (see Figure \ref{fig:example_entity_freq_descriptor_plot} for peak in ``\emph{San Juan}'' and ``\emph{Myrtle Beach}'' mentions).
We defined \emph{pre-peak} as the period that ends $t_{\text{buffer}}$ days before the frequency peak, \emph{during-peak} as the period at most $t_{\text{buffer}}$ days before and at most $t_{\text{buffer}}$ days after, and \emph{post-peak} as the period that begins $t_{\text{buffer}}$ days after the frequency peak (we set $t_{\text{buffer}}=1$).
As described below, we include fixed effects for authors and locations in robustness checks; to improve stability, we removed all locations that are mentioned on fewer than $N=5$ separate dates, and combined all authors with only a single post into a \texttt{RARE} bin.

As shown in the ``\emph{RQ2a (Twitter)}'' column of Table \ref{tab:descriptor_regression_all_results}, the post-peak time period had less descriptor use than the earlier time periods ($\beta=-0.127, p < 0.001$).
Furthermore, descriptor phrase use decreased with the number of days since the start of the event ($\beta=-0.120, p < 0.001$).
These findings are consistent with the hypothesis that entities become more salient through the focus of collective attention, and that this salience makes contextualization less necessary.
The regression also provides more rigorous validation for the trend shown in Figure~\ref{fig:example_entity_freq_descriptor_plot}.

However, an additional potential explanation for the decrease in descriptor context may be a change in the set of authors after the peak in collective attention -- for example, an influx of locals, who are less likely to use descriptors overall.
To test for this, we re-ran the regression above and replaced the author variables (``\emph{local}'' and ``\emph{organization}'') with a fixed effect for each author.
We found that the post-peak effect was still significant and negative ($\beta=-0.253, p < 0.05$). 
This suggests that a change in author population does not explain the decrease in descriptor use over time, or else this would be absorbed by the fixed effects. We note that these findings generally replicate prior work on long-term trends in descriptor phrase usage in non-crisis contexts~\cite{staliunaite2018}, although this prior work did not consider attention ``peak'' as the time variable.

\subsection{Individual Change in Descriptor Context Use}
\label{sec:descriptor_change_individual}
We now further examine temporal dynamics at the level of individual authors (RQ2b).
Under a strong interpretation of our motivating hypotheses, an author who participates frequently in early discussion of the event may use fewer descriptors later during the event, under the assumption that their readers would no longer need context for their event-related posts.
However, other factors may also be at work: an author who has a growing audience may be more likely to use descriptor phrases to accommodate their new readers.

To better model the author-level changes in descriptor use, we introduced the following additional predictors: 
number of prior posts by author during event (author-level), 
number of prior posts by author about the location during event (author-level), 
engagement received by author at $t-1$ (audience),\footnote{We define engagement as the mean of retweets and likes, converted into $z$-scores across the population.} 
and change in engagement received by author between $t-2$ and $t-1$ (audience).
These predictors required a longitudinal sample of frequently-posting authors, i.e. \emph{active} authors, who were identified as those whose post volumes were at or above the $95^{\text{th}}$ percentile among all authors in our collection.
We scraped all publicly available tweets posted by these active authors that mention one of the event's hashtags during the event time period (e.g., all posts for a Harvey-related active author from between August 17 and September 10, 2017 that use \#Harvey or \#HurricaneHarvey).
The locations and descriptor phrases were processed as described in \S~\ref{sec:detect_filter_locations}), and we report the relevant statistics for these active authors in Table~\ref{tab:active_author_data_table}.
We built similar regularized logistic regression models only using data from the active authors who posted at least once during each of the time periods, so as to isolate changes for individual authors.

The results are described in the ``\emph{RQ2b (Twitter})'' column of Table~\ref{tab:descriptor_regression_all_results}.
We find that authors' prior mentions of a location are associated with less descriptor use (${\beta=-0.237}, p < 0.001$) but that there is no significant temporal trend with respect to the start of the event or the peak attention. 
This latter null result held even when we performed the regression without the additional \emph{author} and \emph{audience} variables.
We did find that authors who received more engagement from the audience tended to use more descriptors (${\beta=0.292}, p < 0.001$), which is again consistent with the view that larger audiences necessitate additional contextualization.

We hypothesized that the active authors may be different from the overall population in how they respond to trends in collective attention.
To test this, we identified \emph{regular}  authors as those with lower post volumes below the $95^{\text{th}}$ percentile, and we re-ran the regression analysis with only these individuals.
We found that these less active authors do show a significant decrease in descriptor use following the peak in collective attention (${\beta=-0.127, p < 0.05}$) and a decrease in descriptor use over time (${\beta=-0.098}, p < 0.05$).
It is unclear whether highly active authors have special characteristics, or whether these differences are driven by some other aspect of the design. The set of active authors contains many journalists and news outlets, whose patterns of writing may be shaped by stylistic formalisms but also a greater sensitivity to their audience's awareness of unfolding situations~\cite{murthy2017}.
\section{Discussion}

By examining how people refer to affected locations over the course of crisis events, we found several consistent trends related to audience expectations:
(1) When authors are local to a place, or are writing for an audience who is expected to be local, they are less likely to use descriptor phrases to contextualize references to locations, reflecting shared knowledge among the author and audience;
(2) At a collective level, descriptor use decreases over time during crisis events, even after controlling for a set of explanatory variables;
(3) At an individual level, active authors who receive more prior audience engagement tend to use fewer descriptors, but active authors do not use fewer descriptors over time.
This contrasts with the less-active authors who use fewer descriptors over time.

\subsection{Implications}
\label{subsec:implications}

\paragraph{Theoretical implications for linguistics}
The study highlights uses the unique setting of crisis communication to shed new light on how people present information under varying conditions of shared knowledge~\cite{doyle2015}.
First, when communicating with an audience that is united by geographical affinity, writers tend to omit contextualizing descriptors.
This is predicted by the theory of audience design~\cite{bell1984}, which describes how a speaker modifies their language to fit their expected audience.
Second, we find a distinction between highly active authors and other participants, which complicates the prior understanding of time as a factor in descriptor use: while less active authors tend to decrease the use of descriptors for an entity over time (replicating the findings of \citeauthor{staliunaite2018} \citeyear{staliunaite2018}), more active authors are insensitive to time, but instead vary their writing style in response to audience reaction.
The introduction of these additional predictors gives a more nuanced perspective on how authors anticipate their audience's information needs. 
Rather than viewing collective attention as a single quantity among the public, we argue that attention levels vary across communities, and that individual writers make independent judgments about the degree of attention that is likely to be present in their intended audiences.

\paragraph{Implications for crisis monitoring}

With respect to crisis informatics, this study provides evidence to support theories of information sharing on social media during periods of uncertainty.
Local observers are known to share different types of information during crises, as compared to official organizations~\cite{kogan2015}. 
We find that local observers tend to use fewer contextualizing descriptors, which is consistent with the idea that they are usually writing for an audience of locals who do not need contextualization.
We also find that the use of descriptors decreases after the peak of collective attention during a crisis event. 
This supports previous studies about information sharing during crisis~\cite{houston2015}, which found that people sharing information during a crisis often spend significant time documenting the details of the crisis.
This may mean omitting descriptor phrases (known information) in favor of new details related to the locations affected, which is supported by our finding about URLs.

While we do not offer new tools for crisis informatics, we believe that the findings of this research could help response organizations more effectively track collective attention in crises.
Unlike volume-based metrics, our content-based analysis is relatively robust to problems of missing data, which may occur when internet access is lost during a crisis.
Since organizations often track emergent needs on social media~\cite{imran2015}, highlighting locations that lose descriptor context over time can help organizations better understand public \emph{awareness}, which may be unavailable from more formal accounts of the crisis.
This is especially important in the context of post-crisis recovery~\cite{soden2018} in which organizations often need to assess public awareness of locations in need, because a gap in awareness of a location may result in uncertainty among the public.
Lastly, this study shows that different groups of people (locals, organizations, active authors) use contextualizing information differently, which can shed light on the situation ``on the ground'' more effectively.
Identifying authors who consistently omit context for local entities could help response organizations identify people who are more directly involved with the crisis~\cite{yin2015}, who can then be contacted for further information.

\subsection{Limitations}

\paragraph{Sampling methods}
We focus on only a set of specific crisis events, chosen mainly due to the large volume of online discussions.
It is possible that the patterns observed in our study are specific to these events and locations, so more work is required to establish generalization to other types of crisis events and other types of entities.
Even within these events, our data collection relied on hashtags that may not capture the full breadth of discussion of the crisis events, because we may have missed less frequent hashtags that covered other aspects of the discussion.
If these hashtags are for some reason unrepresentative, then it is possible that a more extensive dataset might reveal other relationships between descriptor use and information expectations, although we have no reason to believe that this is the case.
We focus exclusively on location names because of their geographic relevance to events, but future work should examine other types of named entities (people, organizations) that also undergo change in response to increased attention~\cite{staliunaite2018}.
Finally, we acknowledge that other online contexts might give rise to different expectations about audience knowledge (e.g., discussion forums for news readers versus online encyclopedias whose text is meant to be relevant long after the crisis has passed), and these platforms may therefore feature different patterns of the use of contextualizing descriptors.

\paragraph{Causation and operationalization}
Our findings are strictly correlational: for example, we find that post-peak mentions of locations are less likely to contain descriptor phrases, but given our observational setting, we cannot establish that time is a cause of this linguistic difference. 
We have tried to control for all measurable confounds, but it is always possible that there are unmeasured confounds as well. 
Second, our operationalization of ``peak attention'' is based on the number of mentions during a specific crisis event. 
It is possible that other events attracted attention to the locations under discussion before the crises began, e.g. a political news story relevant to Puerto Rico increased awareness of San Juan before Maria hit, causing a spurious effect.
However, we assume that the hurricanes attracted the most attention to the regions of study, given the massive financial and social impact of the crises.
Third, our operationalization of descriptor phrases is based on a set of lexico-syntactic patterns, using a natural language processing pipeline.
These patterns are not exhaustive, and there may be other syntactic constructions that indicate whether an entity is considered new information for the audience~\cite{rahman2011}.

\subsection{Future work}

Future work should investigate more long-term examples of descriptor use change in news media, including cases where descriptors may reemerge after being dropped as in the case of Flint, Michigan (often referred to as ``\emph{Flint}'' in the early stages of the water crisis that started in 2014).
Identifying common trajectories of descriptor use and writing styles across events can provide insight into how public information needs may shift in response to changing crisis conditions~\cite{olteanu2015}.
With respect to crises, follow-up work should investigate different types of crisis events to determine whether expectations of shared knowledge are significantly different based on the circumstances~\cite{houston2015}.
A fast-moving and highly lethal crisis such as an earthquake may require news media to drop context information quickly to make way for newer or more important information, while a more slow-moving crisis may allow media to retain context to accommodate new readers.
Lastly, future work should consider alternate definitions of ``context'' beyond descriptor phrases.
Longer spans of text may include context information in less direct ways (e.g. ``\emph{San Juan is flooding. It is the capital of Puerto Rico}'') that may still reflect the author's assumed need for context.

\subsection{Conclusion}
This study adds a new content-based perspective to the measurement of collective attention, by analyzing how people discuss breaking news events online.
By examining five recent hurricane events on multiple social media platforms, our research demonstrated how the need for contextual information in location entity references is shaped over time by changing expectations of entity salience.
This study contributes to a better understanding of how people share information during crisis events, and can extend more broadly to other scenarios that involve wide-scale collective attention.
\section*{Acknowledgments}
The authors thank the anonymous reviewers, Sandeep Soni and members of Georgia Tech's SocWeb group for their valuable feedback on earlier drafts.
This research was supported by NSF award IIS-1452443, NIH award R01-GM112697-03. 
DY is supported in part by a grant from Google.

\bibliographystyle{aaai}
\bibliography{main}

\appendix

\section{Detecting author social status}
\label{sec:author_social_status}

In the context of event-based public discussions, it is worth considering whether a post author is (1) local and (2) an organization.
An author who is \emph{local} (more committed) to the event's region will already be aware of the locations under discussion~\cite{kogan2015} and will be less likely to use context than an author who is unfamiliar with the region's locations.
Organizations such as government agencies are often responsible for disseminating official information to help crisis responders and effectively organize aid~\cite{houston2015}.
An author who represents an official organization may want to minimize uncertainty in their messages and use more context than an author who does not represent an organization, i.e. a citizen observer.

We determine author local status and organization status using a sample of metadata available from archived tweets corresponding to the time periods of interest (covering $\sim 20\%$ of all authors in the data).
Following prior work in geolocation (e.g. \citeauthor{kariryaa2018} \citeyear{kariryaa2018}), we approximate the local status of an author posting about an event based on whether the author's self-reported profile location mentions a relevant city or state in the event's affected region (e.g. for Hurricane Maria, a local author would mention ``\emph{Puerto Rico}'' or ``\emph{PR}'' in their location field).

Organizations are difficult to identify automatically, because there is no single indicator of organization status in a Twitter user's profile information.
To determine whether an author counts as an organization, we apply the pre-trained organization classifier from \citeauthor{wooddoughty2018} (\citeyear{wooddoughty2018})\footnote{Accessed 7/2019: \\ \url{https://bitbucket.org/mdredze/demographer/src/peoples2018/}.} to the author's metadata, including name, description, and social attributes.

For both local and organization status, we find reasonable precision with respect to a small subset of hand-labeled authors from our data.\footnote{One of the authors annotated 500 accounts as organizations and locals, based on available metadata, and compared these labels to those produced by the local proxy and organization classifier. The local proxy achieved precision of 87\% and recall of 58\%, and the organization classifier achieved precision of 87\% and recall of 54\%.}

\section{Robustness check for descriptor extraction}
\label{sec:descriptor_robustness_check}

As mentioned in \S \ref{sec:detect_context}, we used the SyntaxNet parser to extract descriptor phrases from the Facebook data due to the parser's better performance on longer sentences.
To verify the consistency of results across parsers, we re-parsed the Facebook data with the \texttt{spacy} parser used for the Twitter and repeated the regression to predict descriptor use from the explanatory factors, i.e. RQ1.
The effect sizes and significance remained the same in the regression on spacy parses, shown in Table \ref{tab:descriptor_regression_only_spacy_results} (cf. ``\emph{RQ1 (Facebook}'' column in Table \ref{tab:descriptor_regression_all_results}).

\begin{table}[]
    \centering
    \begin{tabular}{P{2cm} P{2.5cm} | r r}
         Factor & Variable & Estimate & S.E. \\ \toprule
         Intercept & ~ & -2.08 & 34.4 \\ \midrule
         \textbf{Importance} & Prior location mentions & -0.042 & 8.31 \\ \midrule
         \textbf{Author} & Author in-group posts & -0.172 & 0.549 \\ \midrule
         \textbf{Audience} & Local location & -0.607* & 0.116 \\
         ~ & Group size & 0.106 & 40.3 \\ \midrule
         Deviance & ~ & \multicolumn{2}{r}{469} \\
        Accuracy & ~ & \multicolumn{2}{r}{74.3\%} \\
    \end{tabular}
    \caption{Regression results for Facebook data in RQ1, using \texttt{spacy} parses to detect descriptor phrases. * indicates $p<0.05$, otherwise $p>0.05$ after multiple hypothesis correction.}
    \label{tab:descriptor_regression_only_spacy_results}
\end{table}

\end{document}